\documentclass[conference]{IEEEtran}
\IEEEoverridecommandlockouts
\usepackage{amsmath,amssymb,amsfonts}
\usepackage{algorithmic}
\usepackage{graphicx}
\usepackage[utf8]{inputenc}
\usepackage[english,vietnamese]{babel}
\usepackage[top=1.1in, bottom=1.5in, left=1.1in,right=1.1in]{geometry}
\usepackage{hyperref}
\usepackage{pgfplots}

\begin{document}
\title{A Comparative Study of Neural Network Models for 
  Sentence Classification}  
\author{
\IEEEauthorblockN{Phuong Le-Hong}
\IEEEauthorblockA{Data Science Laboratory\\
College of Science, VNU Hanoi\\
\textit{phuonglh@vnu.edu.vn}}
\and
\IEEEauthorblockN{Anh-Cuong Le}
\IEEEauthorblockA{Faculty of Information Technology\\
  Ton Duc Thang University, Ho Chi Minh City\\
\textit{leanhcuong@tdt.edu.vn}}
}

\maketitle
\selectlanguage{english}
\begin{abstract}
  This paper presents an extensive comparative study of four neural
  network models, including feed-forward networks, convolutional
  networks, recurrent networks and long short-term memory networks, on
  two sentence classification datasets of English and Vietnamese
  text. We show that on the English dataset, the convolutional network
  models without any feature engineering outperform some competitive
  sentence classifiers with rich hand-crafted linguistic features. We
  demonstrate that the GloVe word embeddings are consistently better than
  both Skip-gram word embeddings and word count vectors. We also show the
  superiority of convolutional neural network models on a Vietnamese
  newspaper sentence dataset over strong baseline models. Our
  experimental results suggest some good practices for applying neural
  network models in sentence classification.
\end{abstract}

\section{Introduction}
\label{sec:introduction}

Neural network models have provided a powerful learning method for use in
many natural language problems recently. There are two major types of
neural networks architectures that can be combined in two ways:
feed-forward networks and recurrent networks. While convolutional
feed-forward networks are able to extract local patterns,
recurrent neural networks are able to capture long-range dependency in
the data by abandoning the Markov assumption.

With the emerging interests of the community in deep learning, there
are numerous works in sentence modeling and classification which apply
neural network models. However, to our knowledge, there is not any
attempt to compare these models empirically in sentence
classification, especially in a multilingual setting. In this paper,
we explore different model architectures systematically and
demonstrate that the best performance is obtained by
convolutional neural network models. We compare feed-forward neural
network, recurrent neural networks, and convolutional neural networks
on two datasets: the UIUC question classification dataset for English
and a vnExpress sentence dataset for Vietnamese. 

The main contributions of this paper are as follows. First, we show
that the CNN models without any feature engineering can outperform
some existing competitive question classifiers with rich hand-crafted
linguistic features. Second, we find that the GloVe word vectors are
consistently better than both of the Skip-gram word vectors and word
count vectors when being used in neural network models. Third, we show
the superiority of convolutional neural network models on a Vietnamese
newspaper sentence dataset over strong feed-forward neural network 
models. Finally, these results can serve as a baseline for future
research in these problems.

The remainder of this paper is structured as follows. In
Section~\ref{sec:nnm}, we briefly describe the neural network
architectures in use, including feed-forward networks, convolutional
networks, recurrent networks and its variant long short-term memory
networks. Section~\ref{sec:experiments} presents the experimental
datasets and extensive evaluation
results. Section~\ref{sec:discussion} discusses the results and
related work. Finally, Section~\ref{sec:conclusion} concludes the
paper.

\section{Neural Network Models}
\label{sec:nnm}

\subsection{Feed-Forward Neural Network}

Feed-forward Neural Network (FNN) consists of multiple layers of
nodes. Each layer is fully connected to the next layer in the
network. Nodes in the input layer represent the input data. All other
nodes map inputs to outputs by a linear combination of the inputs with
the node's weights $\mathbf w$ and bias $\mathbf b$ and applying an activation
function. This can be written in matrix form for FNN with $\ell+1$ layers
as follows:
\begin{equation*}
  y(\mathbf x) = f_{\ell}(\cdots f_2(\mathbf w_2^\top f_1(\mathbf w_1^\top
  \mathbf x +
  b_1) + b_2) \cdots + b_{\ell}).
\end{equation*}
Nodes in intermediate layers use logistic function $f(z_i) = 1 / {[1 +
  \exp(-z_i)]}$. Nodes in the output layer use softmax function $f(z_i)
= {\exp(z_i)} / {\sum_{k=1}^K \exp(z_k)}.$
The number of nodes $K$ in the output layer corresponds to the number
of classes. FNN employs backpropagation for learning the model. We
use the logistic loss function for optimization and L-BFGS as an
optimization routine.

\subsection{Convolutional Neural Network}

Convolutional Neural Network (CNN) is a class of FNN which is designed
to require minimal preprocessing. The network learn filters that in
traditional algorithms were hand-engineered. This independence from
prior knowledge and human effort in feature engineering is a major
advantage of CNN.

We build our CNN upon that of ~\cite{Kim:2014} which is originally
proposed for sentence classification. Our CNN consists of six main
layers: (1) a look-up tables to encode words in sentences by their
embeddings, (2) a convolutional layer to recognize $w$-grams, (3) a
non-linear layer with the rectifier activation function, (4) a max
pooling layer to determine the most relevant features, (5) a fully
connected layer with drop-out and (6) a logistic regression layer (a
linear layer with a softmax at the end) to perform classification.

Let $\mathbf s = [w_1, w_2,\dots, w_n]$ be a sentence of length
$n$, where $w_i$ is the $i$-th word of the sentence. Each word $w_i$
is represented by its word embedding $\mathbf x_i$ which is a row
vector of $d$ dimensions. The sentence $\mathbf s$ can now be viewed as
a tensor $\mathbf X = [\mathbf x_1, \mathbf x_2,\dots, \mathbf
x_n]^\top$ of size $n \times d$. 
This matrix is fed into the convolutional layer to extract higher
level features. Given a window size $w$, a filter is seen as a weight
tensor $\mathbf F$ of size $o \times d \times w$, where $o$ is the output
frame size of the filter. The core of this layer is
obtained from the application of the convolutional operator on the two
tensors $\mathbf X$ and $\mathbf F$. The output layer of the
convolutional layer is precisely computed as
\begin{equation*}
  \mathbf Y_{ti} =  \sum_{j=1}^d \sum_{k=1}^w \mathbf F_{ijk} * \mathbf
  X_{t-1+k,j} + b_i, 
\end{equation*}
for all $t = 1, 2,\dots, n-w+1, \forall i=1,2,\dots,o$, 
where $\mathbf b = [b_1,b_2,\dots,b_o]$ is the bias tensor of size
$o$. Then a rectifier linear unit layer is applied element-wise on the
output layer to produce score tensor.

The pooling is then applied to further aggregate the features generated
from the previous layer. The popular aggregating function is $\max$ as
it bears responsibility for identifying the most important
features. More precisely, the max pooling layer produces $\mathbf z =
[z_1,z_2,\dots,z_o]$, where $z_i = \max_{1 \leq t \leq n-w+1} \mathbf
Y_{ti}$. This feature vector is then fed into a fully connected layer of
standard FNN. Following the previous work~\cite{Kim:2014}, we execute a dropout for
regularization by randomly setting to zero a proportion $p$ of the
output elements. Finally, this feature vector is fed into a logistic
regression layer to perform classification.

\subsection{Recurrent Neural Network}

Given an input sequence $[\mathbf x_1, \mathbf x_2,\dots,\mathbf
x_n]$, a standard Recurrent Neural Network (RNN) computes the hidden
vector sequence $[\mathbf h_1, \mathbf h_2,\dots, \mathbf h_n]$ and
outputs vector sequence $[\mathbf y_1, \mathbf y_2,\dots, \mathbf
y_n]$ by iterating the following equations from $t = 1$ to $n$:
\begin{align*}
  \mathbf h_t &= \sigma \left (\mathbf W \mathbf x_t  + \mathbf U
                \mathbf h_{t-1} + \mathbf b^h \right )\\
  \mathbf y_t &= \mathbf V \mathbf h_t + \mathbf b^y
\end{align*}
where $\mathbf W, \mathbf U, \mathbf V$ denote weight matrices (e.g.,
$\mathbf W$ is the input-hidden weight matrix, $\mathbf U$ is the
hidden-hidden weight matrix, and $\mathbf V$ is the hidden-output
weight matrix); the $\mathbf b$ terms denote bias vectors; and
$\sigma$ is the hidden layer function, which is usually an element-wise
application of a sigmoid function.

This simple RNN formulation is sensitive to the ordering of tokens in
the sequence. It was first proposed by Elman~\cite{Elman:90}. Since we
are concerned with the classification problem instead of 
sequence modeling, the hidden vector at the last time step
$\mathbf h_n$ is fed into a fully connected layer with dropout and
then a logistic regression layer to perform classification. 

\subsection{LSTM Network}

In this model, we represent the word sequence of a sentence with a 
LSTM recurrent neural network~\cite{Graves:13}. The LSTM unit at the $t$-th word consists of a
collection of multi-dimensional vectors, including an input gate
$\mathbf i_t$, a forget gate $\mathbf f_t$, an output gate $\mathbf o_t$, a memory cell $\mathbf c_t$,
and a hidden state $\mathbf h_t$. The unit takes as input a $d$-dimensional
input vector $\mathbf x_t$, the previous hidden state $\mathbf h_{t-1}$, the previous
memory cell $\mathbf c_{t-1}$, and calculates the new vectors using the
following six equations:
\begin{align*}
  \mathbf i_t &= \sigma \left (\mathbf W^i  \mathbf x_t + \mathbf U^i
                \mathbf h_{t-1} + \mathbf b^i \right ) \\
  \mathbf f_t &= \sigma \left (\mathbf W^f \mathbf x_t + \mathbf U^f
                \mathbf h_{t-1} + \mathbf b^f \right )  \\
  \mathbf o_t &= \sigma \left (\mathbf W^o \mathbf x_t + \mathbf U^o
                \mathbf h_{t-1} + \mathbf b^o \right )  \\
  \mathbf u_t &= \tanh \left (\mathbf W^u \mathbf x_t + \mathbf U^u
                \mathbf h_{t-1} + \mathbf b^u \right )  \\
  \mathbf c_t &= \mathbf i_t \cdot \mathbf u_t + \mathbf f_t \cdot \mathbf c_{t-1} \\
  \mathbf h_t &= \mathbf o_t \cdot \tanh(\mathbf c_t),
\end{align*}
where $\sigma$ denotes the logistic function, the dot product 
denotes the element-wise multiplication of vectors, $\mathbf W$ and $\mathbf U$ are
weight matrices and $\mathbf b$ are bias vectors. The LSTM unit at $t$-th word
receives the corresponding word embedding as input vector
$\mathbf x_t$.

\section{Experiments}
\label{sec:experiments}

\subsection{Datasets}

We use two datasets in this study. The first one is the UIUC English
question classification dataset.\footnote{Available at
  \href{http://cogcomp.cs.illinois.edu/Data/QA/QC/}{http://cogcomp.cs.illinois.edu/Data/QA/QC/}}
This corpus contains 5,952 manually labeled questions of 6
coarse-grained classes and 50 fine-grained
classes~\cite{Li:2002}. Among them, 500 questions are reserved as the
test set. Question classification is an important task
of question analysis which detects the answer type of the
question. It helps filter out a wide
range of candidate answers and determine answer selection
strategies.~\cite{Le:2014} We report fine-grained
classification accuracy on 50 classes.

The second dataset is a corpus of 20,000 Vietnamese sentences
extracted from the vnExpress online newspaper. Each sentence is
labeled with one of five categories: ``education'', ``entertainment'',
``devices'', ``health'' and ``business''. This dataset is randomly
split into a training set of 16,000 sentences (80\%) and a test set of
4,000 sentences (20\%).

\subsection{Word Embeddings}

The first feature set includes all unigram features which are raw word
tokens. When using neural networks (MLR, FNN, CNN), we transform word
tokens into low-dimensional vectors. In our method, each input word
token is transformed into a vector either by looking up pre-trained
word embeddings or by word hashing with a fixed dimension.

\begin{figure}
  \includegraphics[scale=0.4]{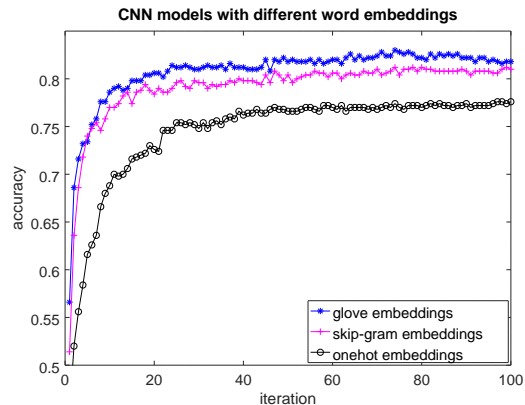}
  \caption{Accuracy of CNN models with different word embeddings,
    including 300-dimension GloVe vectors, 300-dimension Skip-gram
    vectors, and 1,024-dimension one-hot vectors. The maximal length of
  questions is fixed at 20 tokens.}
  \label{fig:cnn-result-6layers}
\end{figure}

For each word token of the input sentence of the UIUC data set, we map to its pre-trained
300-dimensional word vector, either being produced by the Skip-gram
model trained on 3 billion running words of Google News
corpus\footnote{Available at \url{https://code.google.com/archive/p/word2vec/}.} or by the GloVe
model trained on 6 billion running words of Wikipedia 2014 and
Gigaword corpus.\footnote{Available at \url{https://nlp.stanford.edu/projects/glove/}.}. In the word hashing
technique, each word token is mapped to an integer ranging from $0$ to
$d$, where $d$ is the domain dimension. We use the hash function
MurmurHash 3 as feature hashing technique, which is a fast and
space-efficient way of vectorizing features.

Similarly, each word token of the Vietnamese sentence is mapped
to its pre-trained 50-dimensional word vector. These word vectors are
obtained by training a Skip-gram model on a Vietnamese text corpus  of
7.3GB from 2 million articles collected
through a Vietnamese news
portal~\cite{Le:2015c}. Note
that each Vietnamese word may consist of more than one
syllables with spaces in between, which could be regarded as multiple
words by the unsupervised models. Hence it is necessary to replace the
spaces within each word with underscores to create full word
tokens.\footnote{After removal of special characters and tokenization, the articles add
up to $969$ million word tokens.} 

In the following subsections, we first compare the performance of the
models on the English UIUC corpus. We then compare the best CNN models
on the Vietnamese corpus with baseline FNN models.

\subsection{CNN Results}

In the first experiment, we study the effects of two word embeddings
representations, either Skip-gram word vectors or GloVe word vectors,  and
the one-hot encoding representation with dimension
$d=1,024$.\footnote{We also performed experiments with higher dimension
for the one-hot representation but they did not give better
performance.} The layer configuration of the CNNs are kept the same
except the first embedding look-up tables. The convolutional layer has
an output frame size of 256 and kernel width of 3. The non-linear
layer has output size of 128 neurons. The dropout probability is fixed
at $0.1$.


\begin{figure}[t]
  \includegraphics[scale=0.4]{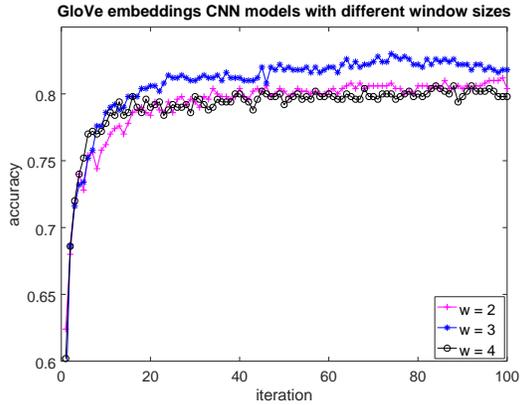}
  \caption{Accuracy of CNN models with different window sizes. The
    configuration of CNN models are kept the same except the window
    size $w$ is varied.}
  \label{fig:cnn-result-6layers-windows}
\end{figure}

This experimental result is shown in
Figure~\ref{fig:cnn-result-6layers}. The $x$-axis is the number of
iterations in training the CNN models. The $y$-axis is accuracy
ratio of the models on the test set. Among the three word
representations, the GloVe representation 
gives the best result. It consistently outperforms the Skip-gram
representation by a clear margin, achieves an accuracy of 83.00\%;
while the Skip-gram representation gives the maximal accuracy of
81.20\%. The one-hot representation has the lowest accuracy, achieving
77.60\%. This experimental result shows the good benefit of word
embeddings learned from large unlabeled text data which capture
syntactic and semantic information.

We also investigate the impact of window size $w$ to the accuracy of CNN
models with GloVe
embeddings. Figure~\ref{fig:cnn-result-6layers-windows} shows the
test accuracy curves with three window sizes of 2, 3, and 4. It is
clear that $w=3$ is the most appropriate window size which gives the
best accuracy.

\subsection{RNN Results}

\begin{figure}
  \centering
  \includegraphics[scale=0.4]{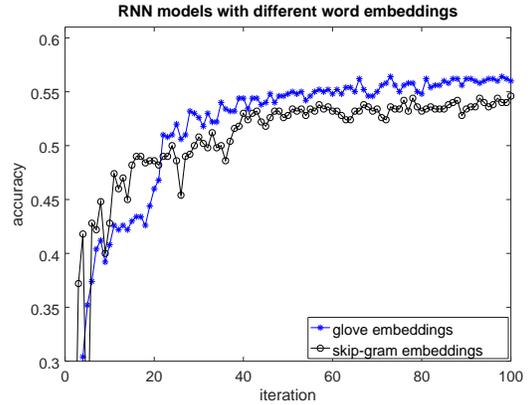}
  \caption{Accuracy of RNN models with two word embeddings
    schemes. The number of hidden units is fixed at 256. The GloVe
    embeddings outperform the Skip-gram embeddings.}
  \label{fig:rnn-result}
\end{figure}

In the second and third experiment, we investigate the performance of
RNN and LSTM models respectively on the two word embedding schemes
GloVe and Skip-gram under the 
same parameter settings.

We first evaluate the performance of RNN models. For each model, we
tune the parameters by grid searching using the test set. The number
of the hidden units in all models is fixed at 256, the batch size is
128, the learning rate is $10^{-2}$, the learning rate decay is $10^{-3}$,
and the optimization algorithm is 
Adagrad. As in previous experiments, we set the iteration number over
the training data as 100.

Figure~\ref{fig:rnn-result} shows the accuracy curves of the two
simple RNN models with the two embeddings schemes. The GloVe
embeddings outperform the Skip-gram embeddings. The best accuracy of
the two RNN models are 56.40\% and 54.60\% respectively. 

\begin{figure}
  \centering
  \includegraphics[scale=0.4]{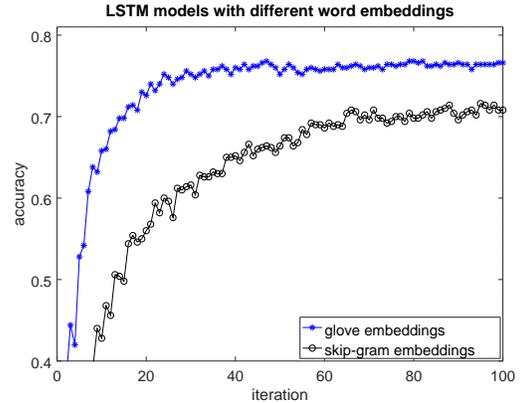}
  \caption{Accuracy of LSTM models with two word embeddings
    schemes. The number of hidden units is fixed at 256. The GloVe
    embeddings outperform significantly the Skip-gram embeddings.}
  \label{fig:lstm-result}
\end{figure}

Figure~\ref{fig:lstm-result} shows the accuracy curves of the two LSTM
models with either GloVe or Skip-gram embeddings. We see that the
GloVe embeddings give significantly better result than the Skip-gram
embeddings. After 100 training iterations, the best accuracy of the
LSTM model with Skip-gram embeddings is only 71.60\% while that of the
LSTM model with GloVe embeddings is 76.80\%. We also see that LSTM
models outperform the simple RNN models by a clear margin but
underperform CNN models. This experimental result demonstrates that
CNN models are better than RNN models in capturing salient features.

\subsection{FNN Results}

In the fourth experiment, we report the performance of FNN
models using count vector representations. The count vector of
a sentence is the common bag-of-word representation of its unigrams
with a minimum frequency cutoff of 2 on the training set. In this representation,
each input sentence is transformed into a count vector of size $d$,
where $d$ is a fixed domain dimension used in the feature hashing
technique. The FNN models use the same
number of 256 units in the hidden layer as in the CNN and RNN
experiments.

\begin{figure}[h]
\begin{center}
  \begin{tikzpicture}
    \begin{axis}[
      height=4cm,
      width=7.5cm,
      symbolic x coords={1024, 2048, 4096, 8192},
      xtick=data,
      ylabel=accuracy,
      nodes near coords={\pgfmathprintnumber[precision=4]{\pgfplotspointmeta}},
      ymin=0.70,ymax=0.8,
      ]
      \addplot[ybar,fill=blue] coordinates {
        (1024, 0.734)
        (2048, 0.746)
        (4096, 0.756)
        (8192, 0.760)
      };
    \end{axis}
  \end{tikzpicture}
\end{center}
\caption{Accuracy of FNN models with one hidden layer of 256 units on
  the UIUC corpus. The $x$-axis 
  shows the count vector encodings with different dimensions, ranging from
  1,024 to 8,196.}
\label{fig:fnn-result}
\end{figure}
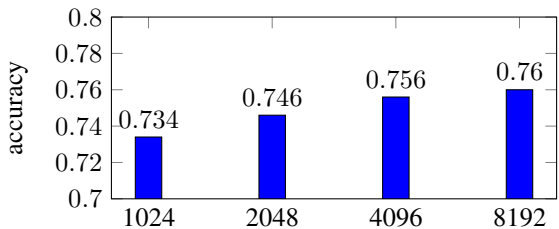

Figure~\ref{fig:fnn-result} shows the accuracy of FNN models. We see
that these models fall behind CNN and RNN models with a large
margin. 

\begin{table}[h]
  \caption{Accuracy scores of CNN, RNN and FNN models on the English
    UIUC question types dataset}
  \begin{center}
    \begin{tabular}{|l|r|}
      \hline
      \textbf{Model}&\textbf{Accuracy}\\
      \hline
      \hline
      CNN with GloVe embeddings & \textbf{83.00}\%\\
      CNN with Skip-gram embedding&81.20\%\\
      CNN with one-hot embeddings&77.60\%\\
      \hline
      RNN with GloVe embeddings&56.40\%\\
      RNN with Skip-gram embeddings&54.60\%\\
      \hline
      LSTM with GloVe embeddings&76.80\%\\
      LSTM with Skip-gram embeddings&71.60\%\\
      \hline
      FNN with bag-of-word vectors&76.00\%\\
      \hline
    \end{tabular}
    \label{tab:uiuc-result}
  \end{center}
\end{table}

Table~\ref{tab:uiuc-result} summarizes the best accuracy scores of the
models. The CNN models are better than the other models by a large
margin. The best model is CNN with GloVe embeddings.

\subsection{vnExpress Results}

\begin{figure}
\begin{center}
  \begin{tikzpicture}
    \begin{axis}[
      height=4cm,
      width=7.5cm,
      symbolic x coords={1024, 2048, 4096, 8192, skip-gram},
      xtick=data,
      ylabel=accuracy,
      nodes near coords={\pgfmathprintnumber[precision=4]{\pgfplotspointmeta}},,
      nodes near coords align={vertical},
      ymin=0.75,ymax=0.8
      ]
      \addplot[ybar,fill=blue] coordinates {
        (1024,   0.761)
        (2048,  0.777)
        (4096,   0.784)
        (8192, 0.787)
        (skip-gram, 0.777)
      };
    \end{axis}
  \end{tikzpicture}
\end{center}
\caption{Accuracy of CNN models on the vnExpress corpus. The $x$-axis
  shows the bag-of-word encodings with different dimensions, ranging from
  1,024 to 8,196 and the Skip-gram word embeddings of 50
  dimensions.}
\label{fig:cnn-result-vne}
\end{figure}
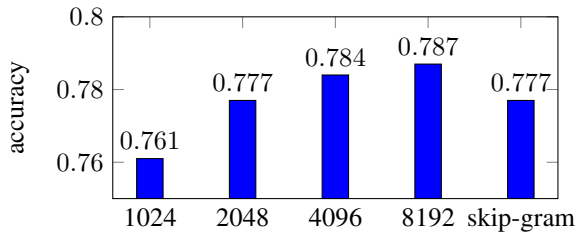

In this subsection, we compare the performance of neural networks
models on the vnExpress corpus. In the fifth experiment, we report
experimental results of the CNN models with different feature
encodings, as shown in the Figure~\ref{fig:cnn-result-vne}. We
see that the word embeddings encoding is slightly outperformed by
bag-of-word encodings with large domain dimensions. However, the training
time of the model with the $8,192$-dimensional bag-of-word encoding is
about four times slower than that of the Skip-gram embeddings.

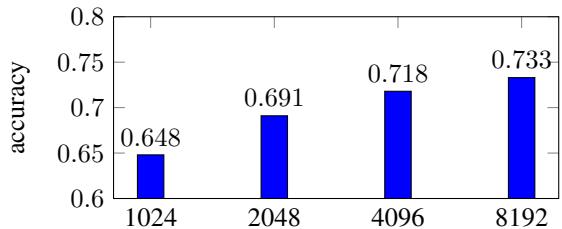
\begin{figure}[h]
\begin{center}
  \begin{tikzpicture}
    \begin{axis}[
      height=4cm,
      width=7.5cm,
      symbolic x coords={1024, 2048, 4096, 8192},
      xtick=data,
      ylabel=accuracy,
      nodes near coords={\pgfmathprintnumber[precision=4]{\pgfplotspointmeta}},
      nodes near coords align={vertical},
      ymin=0.60,ymax=0.8
      ]
      \addplot[ybar,fill=blue] coordinates {
        (1024,   0.648)
        (2048,  0.691)
        (4096,   0.718)
        (8192, 0.733)
      };
    \end{axis}
  \end{tikzpicture}
\end{center}
\caption{Accuracy of FNN models with one hidden layer of 256 units on
  the vnExpress corpus. The $x$-axis 
  shows the bag-of-word encodings with different dimensions, ranging from
  1,024 to 8,196.}
\label{fig:fnn-result-vne}
\end{figure}

Finally, in the sixth experiment, we report the accuracy of the FNN
models on the Vietnamese dataset. In both of the CNN and the FNN models, we do
not tune them for their best performance but intentionally use a
fully-connected hidden layer of the same 256 hidden units. With this setting,
their salient feature detection capability can be directly
comparable. Figure~\ref{fig:fnn-result-vne} shows the result. We see
that the best FNN model is worse than all CNN models: the accuracy
gap between the two best models is about 6\% of absolute points.

\section{Discussion}
\label{sec:discussion}

RNN models have been shown to be a very strong
sequence learner in that it can detect intricate patterns in the data
and long-range dependency. However, we show that this power is not
needed for sentence classification in both the UIUC question dataset
for English and the vnExpress sentence dataset for Vietnamese. The
word order and sentence structure are not really important in
these cases. The bag-of-word or bag-of-ngram classifier just work as
well or even better than RNN models, including the powerful LSTM
models.

The CNN models for sequence learner have been designed to identify
indicative local features in a long sequence and to combine them. They
are able to capture $n$-grams that are predictive for sentence
classification, without the need to specify a very sparse vector for
each possible $n$-gram as in the traditional bag-of-ngram approach. As
a result, CNN models are not only effective -- avoiding data
sparsity problems, but also scalable -- any window size produces a
fixed size vector representation of the sentence. Our experiments have
demonstrated that CNN models outperform all other strong competitive models on
two sentence datasets of different natural languages.

In our experiments, we do not use any feature engineering, only raw
sentences are provided. All the models only use either word identity
or pre-trained word embeddings for the concerned languages
(300-dimensional Skip-gram word vectors, 300-dimensional GloVe word
vectors for English, and 50-dimensional Skip-gram word vectors for
Vietnamese). The models can be thus considered
language-independent.

In particular, on the UIUC question dataset, Li and
Roth~\cite{Li:2002} developed the first machine learning approach to
question classification which uses the SNoW learning
architecture. Using the feature set of lexical words, part-of-speech
tags, chunks and named entities, they achieved 78.8\% of fine-grained
accuracy. The UIUC dataset has inspired many follow-up works on
question classification. Zhang and Lee~\cite{Zhang:2003} used linear
support vector machines (SVM) with all question $n$-grams and obtained
79.2\% of accuracy. Hacioglu and Ward~\cite{Hacioglu:2003} used linear
SVM with question bigrams and error-correcting codes and achieved
82.0\% of accuracy. Our CNN models with GloVe embeddings can achieve
83.00\% of fine-grained accuracy, which are better than some early
question classifiers.\footnote{Some recent question classifiers have
  integrated head words, their hypernyms and other semantic features
  to obtain an accuracy of about
  91\%. See~~\cite{Li:2006,NguyenLeMinh:2007,Huang:2008,NguyenVanTu:2016,DangHaiTran:2013},
  for more detail.}


\section{Conclusion}
\label{sec:conclusion}

In this paper, we compare different neural network models for sentence
classification, including FNN, RNN, LSTM, and CNN networks.
In these models, features are automatically learned without any
complicated natural language processing. Experimental results on
two sentence datasets, one for English and one for Vietnamese show
that the CNN models significantly outperform other models on both of
the datasets. In particular, on the UIUC English question
classification dataset, the GloVe embeddings are consistently better
than the Skip-gram embeddings. On this dataset, our CNN models without
any feature engineering also outperform some existing question
classifiers with rich hand-crafted linguistic features.



\bibliographystyle{IEEEtran}
\bibliography{references}
\end{document}